\documentclass{article}

\PassOptionsToPackage{numbers,compress}{natbib}
\usepackage[preprint]{neurips_2026}
\usepackage{amsmath}
\usepackage{graphicx}
\usepackage{comment}
\graphicspath{
{./disentanglement_top1_heatmap_final_paper_use/}
{./cka_top1_heatmap_final_paper_use/}
}

\graphicspath{
{./disentanglement_top5_heatmap_final_paper_use/}
{./cka_top5_heatmap_final_paper_use/}
}

\usepackage[utf8]{inputenc} 
\usepackage[T1]{fontenc}    
\usepackage{hyperref}       
\usepackage{url}            
\usepackage{booktabs}       
\usepackage{amsfonts}       
\usepackage{nicefrac}       
\usepackage{microtype}      
\usepackage{xcolor}         
\usepackage{amssymb}
\usepackage{microtype}
\usepackage{subcaption}

\title{What Does the Brain See? Multiview Neural Representations to Demystify the Brain-Visual Alignment}

\author{%
  Salini Yadav~\thanks{Corresponding author. Email: \texttt{salini\_y1@cs.iitr.ac.in}} \\
  Department of CSE\\
  IIT Roorkee, India \\
  \And
  Taveena Lotey \\
  Department of CSE\\
  IIT Roorkee, India \\
  \AND
  Pravendra Singh \\
  Department of CSE\\
  IIT Roorkee, India \\
  \And
  Partha Pratim Roy \\
  Department of CSE\\
  IIT (ISM) Dhanbad, India
  }

\begin{document}

\maketitle


\begin{abstract}
Zero-shot visual decoding from electroencephalography (EEG) aims to infer visual semantics from non-invasive neural recordings, but remains challenging due to the low signal-to-noise ratio, non-stationarity, and limited spatial resolution of EEG. 
Existing EEG--vision alignment methods often rely on holistic EEG embeddings, which can obscure the complementary temporal, spectral, and spatial structure underlying visual perception. 
We introduce a unified multiview EEG representation learning framework for aligning brain responses with visual semantic embeddings. 
Our method builds an EEG encoder that jointly models three complementary views: input-conditioned state-space temporal dynamics, learnable wavelet-based spectral decomposition for sample-adaptive frequency modeling, and attention-modulated graph learning for structured electrode interactions. 
The resulting multiview EEG embeddings are fused and aligned with pretrained visual representations in a shared semantic space using contrastive learning with EEG-specific regularization, enabling 200-way zero-shot visual classification. 
Experiments on THINGS-EEG benchmark show that our method achieves state-of-the-art performance, with \textbf{54.8\% Top-1 and 85.6\% Top-5 accuracy} in the within-subject setting and \textbf{15.3\% Top-1 and 45.4\% Top-5 accuracy} in the cross-subject setting. 
We further present the first systematic cross-session EEG--image decoding evaluation, achieving \textbf{40.8\% Top-1 and 78.0\% Top-5 accuracy}.
These results suggest that explicitly modeling multiview neural structure improves both semantic alignment and generalization in EEG-based visual decoding.
\end{abstract}

\section{Introduction}

Humans recognize objects in complex visual scenes rapidly and reliably, transforming sensory input into semantic representations that support perception, decision-making, and interaction \cite{biederman1987recognition,goodale1992separate,tanaka1996inferotemporal}. 
Decoding such visual representations from brain activity is a central goal of cognitive neuroscience and brain--computer interface (BCI) research \cite{mridha2021brain}. 
While fMRI has enabled high-resolution visual reconstruction and categorization \cite{kay2008identifying,nishimoto2011reconstructing}, its slow hemodynamic response limits temporal precision. 
MEG provides richer temporal information \cite{cichy2014resolving,tong2012decoding}, but remains costly and difficult to deploy. 
Electroencephalography (EEG), in contrast, is inexpensive, portable, and provides millisecond-level temporal resolution, making it attractive for scalable visual decoding \cite{li2024noninvasive}. 
However, EEG signals are noisy, non-stationary, and spatially coarse, making semantic visual decoding substantially more challenging \cite{spampinato2017deep,bashivan2015learning}.

Recent large-scale datasets such as THINGS-EEG have expanded EEG-based visual decoding to thousands of natural images and diverse object concepts \cite{hebart2019things,gifford2022large}. 
Meanwhile, self-supervised and multimodal vision models, especially CLIP, provide semantic embedding spaces that support open-vocabulary and zero-shot recognition \cite{oord2018representation,chen2020simple,radford2021learning}. 
These developments have motivated EEG--vision alignment methods that map neural signals into pretrained visual-semantic spaces for zero-shot classification or image reconstruction \cite{ferrante2024decoding}. 
Nevertheless, their performance is often limited by the quality and stability of the EEG representations being aligned.

A key limitation of existing EEG--vision alignment approaches is their reliance on holistic EEG embeddings. 
Such embeddings may capture discriminative information, but they often overlook the heterogeneous structure of EEG responses. 
Visual EEG signals contain temporal dynamics reflecting evolving perceptual processing, spectral patterns associated with frequency-specific neural activity, and spatial dependencies induced by electrode topology and cortical organization. 
Prior work has modeled these dimensions separately or through partial interactions \cite{zhang2025category,sun2025spatial,mai2024learning,10198371,hameed2024temporal}. 
However, a unified framework that jointly captures long-range temporal dependencies, adaptive frequency structure, and input-dependent spatial interactions remains underexplored. 
This fragmentation can produce unstable EEG embeddings that align weakly with visual semantics, particularly under cross-subject and cross-session evaluation.

In this paper, we propose a unified multiview EEG representation learning framework for zero-shot visual decoding. 
Rather than modifying the pretrained vision encoder, we focus on learning a stronger EEG encoder that produces stable and semantically aligned neural embeddings. 
Our framework models three complementary views of EEG: an input-conditioned state-space temporal module for long-range and non-stationary dynamics, a learnable wavelet-based spectral module for sample-adaptive frequency decomposition, and an attention-modulated graph module for structured spatial interactions among electrodes. 
The resulting temporal, spectral, and spatial features are fused into compact EEG embeddings and aligned with pretrained visual representations using contrastive learning and EEG-specific regularization.

We evaluate our method on THINGS-EEG benchmark under within-subject, cross-subject, and cross-session settings. 
In addition to standard within-subject and cross-subject evaluation, we present the first systematic cross-session EEG--image decoding evaluation to assess whether semantic alignment remains stable across recording sessions. 
Across these settings, our approach improves 200-way zero-shot visual classification, indicating that explicit multiview modeling provides a more reliable basis for EEG--visual semantic alignment.

Our main contributions are summarized as follows:
\vspace{-0.3em}
\begin{itemize}
    \item We introduce a unified multiview EEG representation learning framework for zero-shot visual decoding, jointly modeling temporal dynamics, adaptive spectral structure, and spatial electrode interactions.

    \item We design a complementary EEG encoder with an input-conditioned state-space temporal module, a learnable wavelet-based spectral module, and an attention-modulated graph spatial module.

    \item We incorporate EEG-specific regularization to improve representation stability through trial-level consistency and class-level semantic alignment.

    \item We achieve state-of-the-art performance on THINGS-EEG for 200-way zero-shot visual classification and present a systematic cross-session EEG--image decoding evaluation.
\end{itemize}

\section{Related work}
\label{gen_inst}

Recent progress in visual neural decoding has been accelerated by large-scale datasets such as THINGS-EEG \cite{gifford2022large} and by pretrained multimodal and generative models. 
Existing work mainly focuses on zero-shot semantic decoding, generative reconstruction, and reducing the gap between visual stimuli, neural responses, and semantic representations.

\subsection{Zero-shot visual decoding frameworks}

Zero-shot visual decoding aims to infer visual semantics beyond categories observed during training. 
Recent methods commonly align neural signals with pretrained semantic spaces, especially CLIP embeddings. 
NICE-EEG \cite{songdecoding} introduced hierarchical contrastive learning for EEG--CLIP alignment on THINGS-EEG, while NICE-LLM \cite{song2025recognizing} used LLM-generated captions to strengthen semantic supervision. 
BraVL \cite{du2023decoding} modeled brain, vision, and language jointly, showing the value of linguistic priors for semantic generalization. 
UMind \cite{xu2025umind} further unified retrieval, classification, and reconstruction using coarse- and fine-grained textual descriptions. 
Although these methods improve semantic alignment, they often rely on relatively simple EEG encoders and leave the temporal, spectral, and spatial structure of EEG representations insufficiently explored.

\subsection{Generative reconstruction with diffusion models}

Beyond retrieval and classification, neural decoding has also been formulated as a conditional image generation problem. 
Early reconstruction methods based on GANs were limited by instability during training and high data requirements, motivating the use of pretrained diffusion models. 
Takagi and Nishimoto \cite{takagi2023high} showed that latent diffusion models can reconstruct high-resolution images from fMRI signals. 
This direction was later extended to EEG, where DreamDiffusion \cite{bai2023dreamdiffusion} generated plausible images from noisy EEG inputs, and ATM-S \cite{li2024visual} improved reconstruction by mapping EEG features into CLIP's latent space. 
While diffusion-based approaches improve visual quality, their success still depends on the stability and semantic fidelity of the EEG embeddings used as conditioning signals.

\subsection{Bridging the semantic and modality gap}

Several recent works address the mismatch between visual stimuli, perceptual representations, and neural recordings. 
UBP \cite{wu2025bridging} introduced an uncertainty-aware blur prior to reduce perceptual information loss and measurement noise. 
Human-Aligned Image Models \cite{rajabi2025human} showed that vision encoders aligned with human perceptual judgments improve decoding across EEG and fMRI. 
CognitionCapturer \cite{zhang2025cognitioncapturer} incorporated multimodal priors, including depth and textual representations, to better capture cognitive states underlying visual perception. 

These works highlight the importance of semantic and perceptual alignment, but many still assume fixed or weakly structured EEG encoders. 
In contrast, our method focuses on improving the EEG representation itself by jointly modeling long-range temporal dynamics, adaptive spectral structure, and spatial electrode interactions, enabling more stable alignment with pretrained visual representations.

\section{Method}

We present a multiview EEG--image alignment framework for zero-shot visual decoding. 
The following sections define the problem formulation, describe the proposed multiview EEG--image decoding framework, and introduce the training objective.

\subsection{Problem formulation}

Let $\mathbf{X} \in \mathbb{R}^{C \times T}$ denote a multi-channel EEG signal with $C$ electrodes and $T$ time points, recorded while a subject views a visual stimulus $I$.
Our objective is to learn an EEG encoder $f_{\theta}(\cdot)$ that maps neural signals to a semantic embedding
$\mathbf{z}_{\mathrm{eeg}} \in \mathbb{R}^{d}$ aligned with a pretrained vision embedding $\mathbf{z}_{\mathrm{img}} \in \mathbb{R}^{d}$, where $d$ is the embedding dimension. By learning this alignment, visual decoding can be formulated as zero-shot semantic classification in a shared embedding space, enabling generalization beyond closed-set categories. Given a dataset of paired EEG recordings and visual stimuli, the encoder is trained to produce discriminative and semantically meaningful EEG representations that reflect the underlying perceptual content.


\begin{figure}
  \centering
  \includegraphics[width=\linewidth]{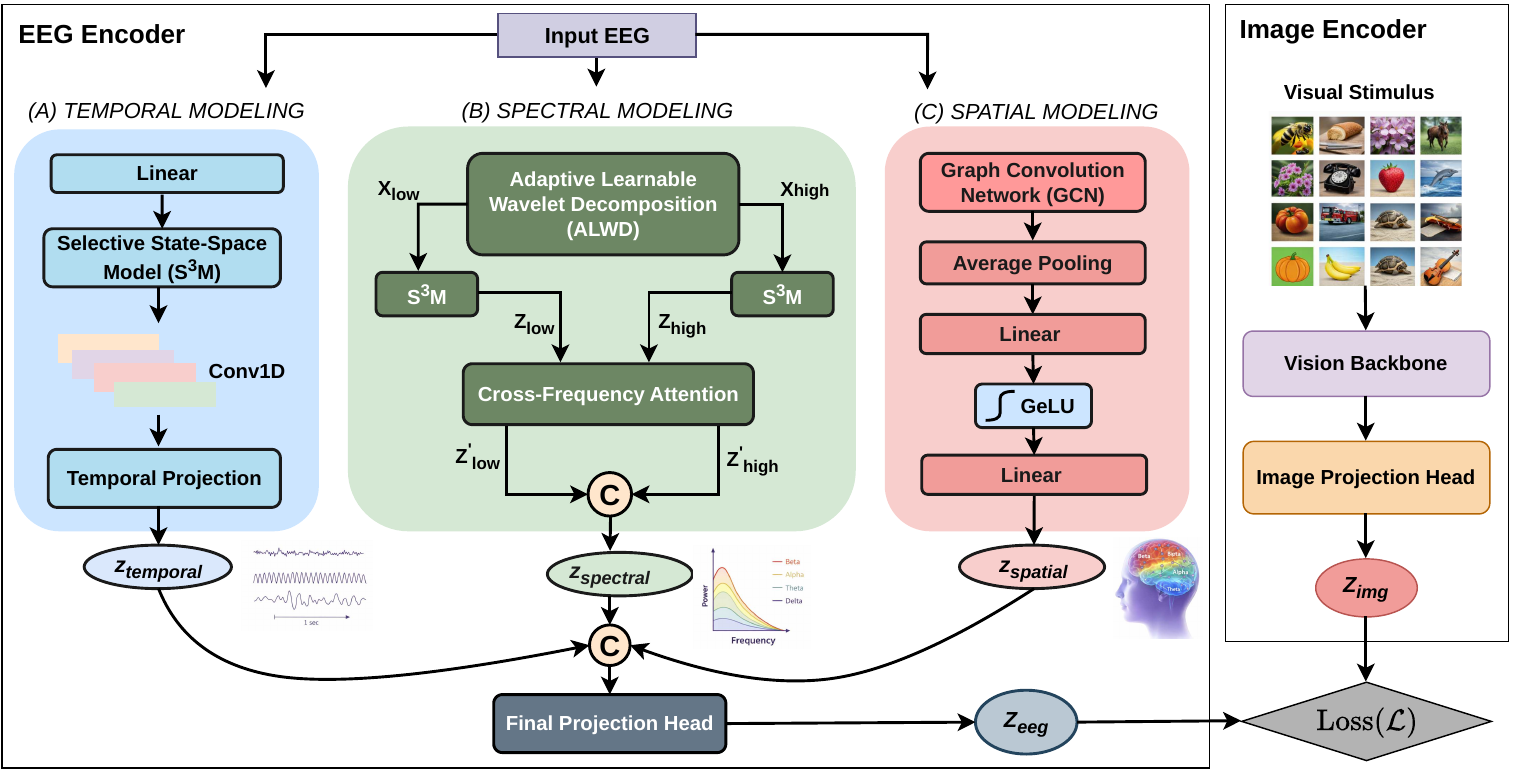}
  \caption{Overview of the proposed multiview EEG–Image decoding framework. Given a multichannel EEG trial, the EEG encoder constructs complementary temporal, spectral, and spatial representations using selective state-space modeling ($S^{3}\mathrm{M}$), adaptive learnable wavelet-inspired decomposition (ALWD), and an attention-modulated graph convolutional network (GCN). These representations are fused into a compact EEG embedding, which is aligned with pretrained visual representations for zero-shot semantic decoding.}
\label{fig:architecture}
\vspace{-0.9mm}
\end{figure}


\subsection{Multiview EEG--Image decoding framework}
As illustrated in Figure~\ref{fig:architecture}, the proposed framework is a multimodal architecture that aligns EEG signals with visual stimuli. It consists of a dual-stream setup: an EEG encoder that factorizes input signals into complementary temporal, spectral, and spatial components, and a pretrained image encoder that extracts visual representations. The EEG components are processed in parallel and then fused into a unified embedding. Finally, both the EEG and image features are projected into a shared latent space and aligned using a contrastive alignment loss, augmented with EEG-specific regularization terms.

\subsubsection{\textbf{EEG ENCODER}}

The key blocks of the EEG encoder are (a) Temporal Modeling, (b) Spectral Modeling, and (c) Spatial Modeling. These blocks are discussed in detail as follows.

\textit{(A) TEMPORAL MODELING }

To capture long-range and input-adaptive temporal dependencies in EEG signals, we employ a Selective State-Space Model (S$^{3}$M). Given an input EEG sequence $X \in \mathbb{R}^{C \times T}$, the model maintains a latent state $h_t \in \mathbb{R}^{d_s}$ that evolves over time according to
\vspace{-0.5em}
\begin{equation}
    h_t = \bar{A}_t \odot h_{t-1} + \bar{B}_t x_t,
    \qquad
    \bar{A}_t = \exp(\Delta_t A),
\end{equation}
where $x_t$ is the embedded EEG token at time step $t$, $\Delta_t$ is an input-dependent step size, $A$ is a learnable state transition parameter, and $\bar{B}_t$ is an input-dependent input projection.
The output at each time step is computed as
\vspace{-0.5em}
\begin{equation}
    y_t = \textstyle \sum_{k=1}^{d_s} C_{t,k} \, h_{t,k}
\end{equation}
where $C_t \in \mathbb{R}^{d_s}$ is an input-dependent readout vector.
The sequence of outputs from the $S^{3}\mathrm{M}$ is subsequently processed with a 1D convolution (Conv1d) along the temporal axis to capture local temporal patterns and refine short-range dependencies. The convolved features are then passed through a temporal projection layer that maps the representations to the desired feature dimensionality. The resulting representation ($z_{\text{temporal}}$) serves as the final temporal feature embedding for subsequent processing.

\textit{(B) SPECTRAL MODELING }
\label{subsec:spectral_dynamics}

EEG signals exhibit complex temporal-spectral characteristics, in which discriminative information may lie in frequency bands that vary across subjects and tasks. Conventional EEG analysis pipelines rely on predefined frequency bands (e.g., delta, theta, alpha), which impose rigid and task-agnostic assumptions. To overcome these limitations, we introduce an \emph{adaptive learnable wavelet decomposition} (ALWD) that dynamically selects frequency representations conditioned on the input EEG signal.

Let $\{\phi_k, \psi_k\}_{k=1}^K$ denote a bank of learnable low-pass and high-pass filters. We compute sample-specific selection weights $\mathbf{w} = \text{Softmax}(\mathcal{M}(X)) \in \mathbb{R}^K$ via a projection network $\mathcal{M}$. The input-adapted filters ($\Phi, \Psi$) are synthesized as weighted combinations, and the spectral components are obtained via strided convolution ($\circledast_{\downarrow 2}$):
\begin{equation}
\label{eq:alwd}
\begin{gathered}
    \Phi = \textstyle \sum_{k=1}^K w_k \phi_k, \qquad \Psi = \textstyle \sum_{k=1}^K w_k \psi_k, \\
    X_{low} = X \circledast_{\downarrow 2} \Phi, \qquad X_{high} = X \circledast_{\downarrow 2} \Psi.
\end{gathered}
\end{equation}
\noindent This process produces complementary low- and high-frequency EEG representations with distinct temporal characteristics. Specifically, $X_{low}$ captures slow, stable neural oscillations, while $X_{high}$ emphasizes transient, rapidly changing activity.
To account for these differences, we apply separate Selective State-Space Models ( $S^{3}\mathrm{M}$) to each spectral component, as defined in Eq.~(\ref{eq:ssm_spectral}).
\vspace{-0.2em}
\begin{equation}
\label{eq:ssm_spectral}
\begin{split}
    z_{low} &= \mathrm{S^{3}M}_{low}(X_{low}), \\
    z_{high} &= \mathrm{S^{3}M}_{high}(X_{high}).
\end{split}
\end{equation}
By conditioning state transitions on the input signal, the  $S^{3}\mathrm{M}$  naturally adapts its effective temporal memory to the spectral content of each component, enabling slow state evolution for low-frequency dynamics and faster adaptation for high-frequency transients.

To capture interactions across frequency scales, we further employ cross-frequency attention, allowing information exchange between low- and high-frequency temporal representations. Let $\mathrm{Attn}(Q, K, V)$ denote the standard multi-head attention operation. The cross-attended features are computed as:
\vspace{-0.3em}
\begin{equation}
\label{eq:cross_attn}
\begin{split}
    z'_{low} &= \mathrm{Attn}(z_{low}, z_{high}, z_{high}), \\
    z'_{high} &= \mathrm{Attn}(z_{high}, z_{low}, z_{low}).
\end{split}
\end{equation}
\noindent In this formulation, $z'_{low}$ integrates high-frequency context into the stable low-frequency features, while $z'_{high}$ incorporates low-frequency trends into the transient high-frequency features. The attended features $z^{'}_{low}$ and $z^{'}_{high}$ are then concatenated, temporally upsampled to the original resolution, and projected to obtain the spectral embedding $z_{spectral} \in \mathbb{R}^{ B\times d}$.


\textit{(C) SPATIAL MODELING }
\label{subsec:electrode_graph}

EEG electrodes are spatially distributed over the scalp, capturing neural activity from interacting brain regions. Consequently, meaningful EEG representations must account for spatial dependencies rather than treating electrodes independently. To this end, we employ a Graph Convolutional Network (GCN) in which electrodes serve as nodes. We define a dynamic adjacency matrix $\mathcal{A}$ that combines a learnable global structure $A_{global}$ with input-dependent edge attention $A_{att}(X)$. The spatial embedding ($z_{spatial}$) is computed as:
\vspace{-0.5em}
\begin{equation}
\label{eq:spatial_gcn}
\begin{gathered}
    \mathcal{A} = A_{global} \odot A_{att}(X), \\
    z_{spatial} = \sigma(\mathcal{A} X W_{GCN}).
\end{gathered}
\end{equation}
\noindent where $W_{GCN}$ is the weight matrix and $\sigma$ denotes the non-linear activation. This formulation captures both shared spatial structure and sample-specific functional interactions.

Finally, the EEG encoder integrates complementary representations that capture temporal, spectral, and spatial information. These embeddings are concatenated and mapped through a final projection $W_{proj}$ to yield the aligned EEG representation $z_{eeg}$:
\vspace{-0.5em}
\begin{equation}
\label{eq:final_fusion}
    z_{eeg} = W_{proj} [z_{temporal} \parallel z_{spectral} \parallel z_{spatial}]
\end{equation}
\noindent where $\parallel$ denotes the concatenation operation.

\subsubsection{\textbf{IMAGE ENCODER}}

In this work, we employ CLIP-ViT-H-14 \citep{Radford2021LearningTV} as the image encoder, pretrained on large-scale image datasets. The encoder extracts semantically discriminative visual embeddings that serve as fixed targets for contrastive alignment with EEG representations. To this end, all images are resized to $224 \times 224$ to ensure consistent and stable feature representations. All stimulus images are passed through the pretrained CLIP-ViT-H-14 once prior to training the EEG model, and the resulting visual embeddings are stored and reused throughout training. The image encoder parameters are frozen during training, which reduces computational cost and ensures a stable visual embedding space for learning EEG--image correspondence. The image encoder produces an image representation $z_{\text{img}}$, which is then used for cross-modal alignment.

\subsubsection{\textbf{Training objective}}

We aim to learn a shared embedding space that aligns neural responses with visual stimuli.
Let $\mathbf{z}_{\text{eeg}}, \mathbf{z}_{\text{img}} \in \mathbb{R}^d$ denote the $\ell_2$-normalized EEG and image embeddings, respectively.
We optimize the symmetric contrastive alignment loss as follows:
\vspace{-0.5em}
\begin{equation}
\mathcal{L}_{\text{align}}
=
-\frac{1}{B}
\sum_{i=1}^{B}
\log
\frac{
\exp(\langle \mathbf{z}_{\text{eeg}}^{(i)}, \mathbf{z}_{\text{img}}^{(i)} \rangle / \tau)
}{
\sum_{j=1}^{B}
\exp(\langle \mathbf{z}_{\text{eeg}}^{(i)}, \mathbf{z}_{\text{img}}^{(j)} \rangle / \tau)
},
\end{equation}
\noindent where $\tau$ is a temperature parameter.
To account for EEG variability and class-level structure, we introduce repetition consistency regularization ($\mathcal{L}_{\text{RCR}}$) and prototype-level contrastive regularization ($\mathcal{L}_{\text{CPCA}}$). These regularizers introduce EEG-specific inductive biases that improve robustness to trial-level variability and enhance class-level semantic structure. 
The total training objective is:
\vspace{-0.5em}
\begin{equation}
\mathcal{L}
=
\mathcal{L}_{\text{align}}
+
\lambda_{\text{RCR}} \mathcal{L}_{\text{RCR}}
+
\lambda_{\text{CPCA}} \mathcal{L}_{\text{CPCA}},
\end{equation}
\noindent where $\lambda_{\text{RCR}}$ and $\lambda_{\text{CPCA}}$ control the regularization strengths.

\vspace{-0.3em}
\section{Experimental setup}
\label{others}

We evaluate the proposed framework on THINGS-EEG benchmark under within-subject, cross-subject, and cross-session zero-shot decoding settings. 
This section summarizes the dataset, evaluation protocols, and implementation details.

\subsection{Dataset and preprocessing}
We evaluate our method on \textbf{THINGS-EEG} dataset \cite{gifford2022large}, a large-scale benchmark for EEG-based visual object recognition. The dataset contains EEG recordings collected while participants viewed natural images drawn from THINGS-EEG database, which includes 1,854 object concepts. EEG data were recorded from 10 subjects, each completing four sessions. The training set consists of 16,540 image conditions (1,654 concepts $\times$ 10 images), each repeated four times, while the test set contains 200 held-out object concepts, each presented 80 times, ensuring no overlap in object concepts between training and testing. 

\subsection{Task definitions and evaluation protocols}




We evaluate EEG-image decoding under three zero-shot settings: \textbf{within-subject}, \textbf{cross-subject}, and \textbf{cross-session}, where test classes are disjoint from training classes.

In the within-subject setting, models are trained and evaluated separately for each subject using subject-specific splits. In the cross-subject setting, models are trained on multiple subjects and evaluated on a held-out subject to assess cross-subject generalization. 
In the cross-session setting, for each subject, models are trained on EEG trials from three recording sessions and evaluated on the held-out fourth session. 
This protocol tests robustness to session-dependent distribution shifts caused by changes in recording noise, electrode contact, and subject state.
Although cross-session evaluation has been studied in other EEG tasks, it remains largely unexplored in the context of EEG-image semantic decoding.

Zero-shot classification is performed by aligning EEG embeddings with image-based class prototypes in a shared semantic space. Performance is evaluated using Top-1 and Top-5 accuracy, where Top-1 measures exact prediction correctness and Top-5 checks whether the correct class appears among the five most confident predictions.

\subsection{Implementation details}


All models are trained for 50 epochs using the AdamW optimizer with a learning rate of $1\times10^{-4}$ and weight decay of $1\times10^{-4}$. 
Following common practice in THINGS-EEG evaluation, we report results from a fixed random seed to ensure direct comparability with prior work.
We use a batch size of 64 for training, validation, and testing. All experiments are conducted on a single NVIDIA RTX 4500 Ada GPU with 24~GB memory. 

\section{Results}
This section presents a comprehensive evaluation of the proposed method under three zero-shot EEG decoding settings: \emph{within-subject}, \emph{cross-subject}, and \emph{cross-session} classification. We compare against recent representative methods, including BraVL~\cite{du2023decoding}, NICE, NICE-SA, NICE-GA~\cite{songdecoding}, ATM-S~\cite{li2024visual}, UMind~\cite{xu2025umind}, CognitionCapturer (CC)~\cite{zhang2025cognitioncapturer}, and UBP~\cite{wu2025bridging}. Across all settings, the proposed method consistently achieves superior performance, demonstrating stronger robustness and semantic alignment of learned representations. 

\subsection{Within-Subject zero-shot classification}
\begin{figure}[!t]
  \centering
  \begin{subfigure}{0.48\linewidth}
    \centering
    \includegraphics[width=\linewidth]{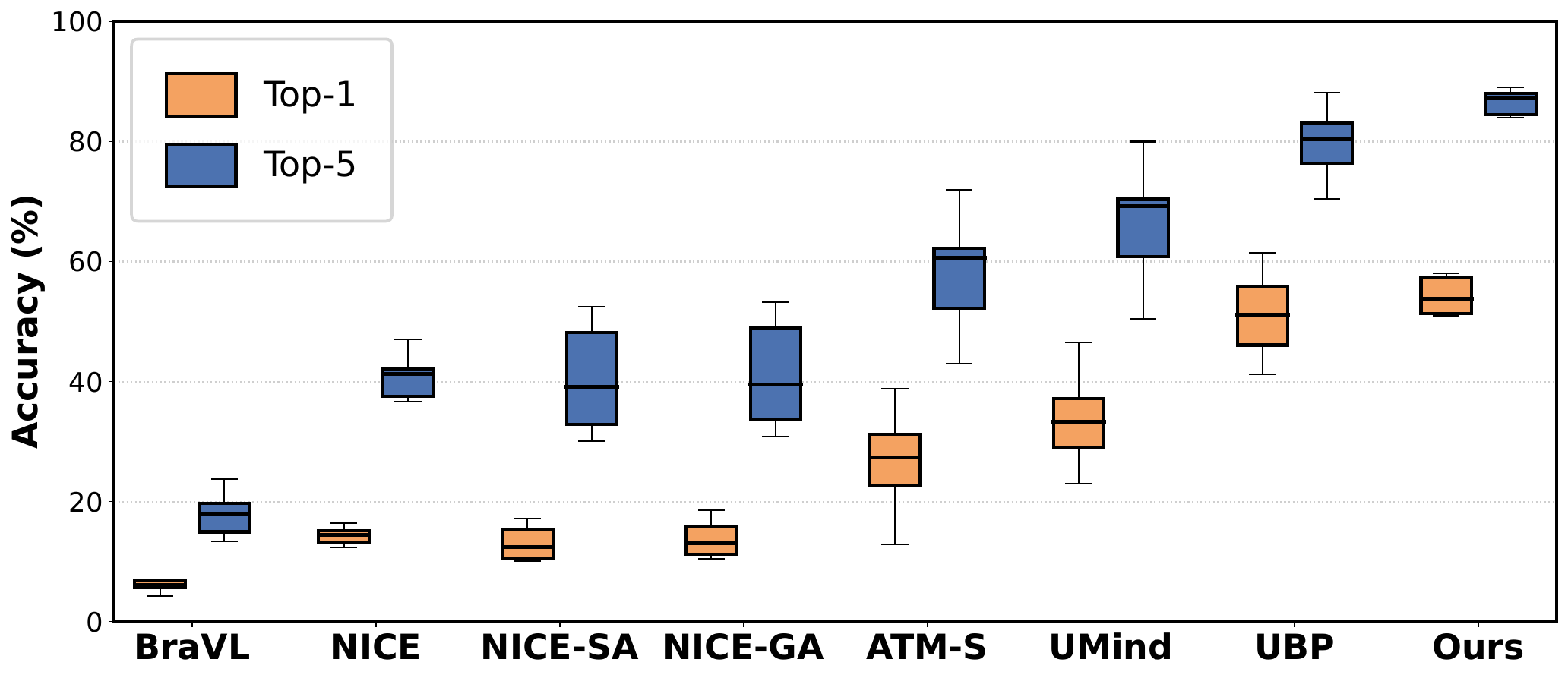}
    \caption{Within-subject}
    \label{fig:boxplots_within}
  \end{subfigure}
  \hfill
  \begin{subfigure}{0.48\linewidth}
    \centering
    \includegraphics[width=\linewidth]{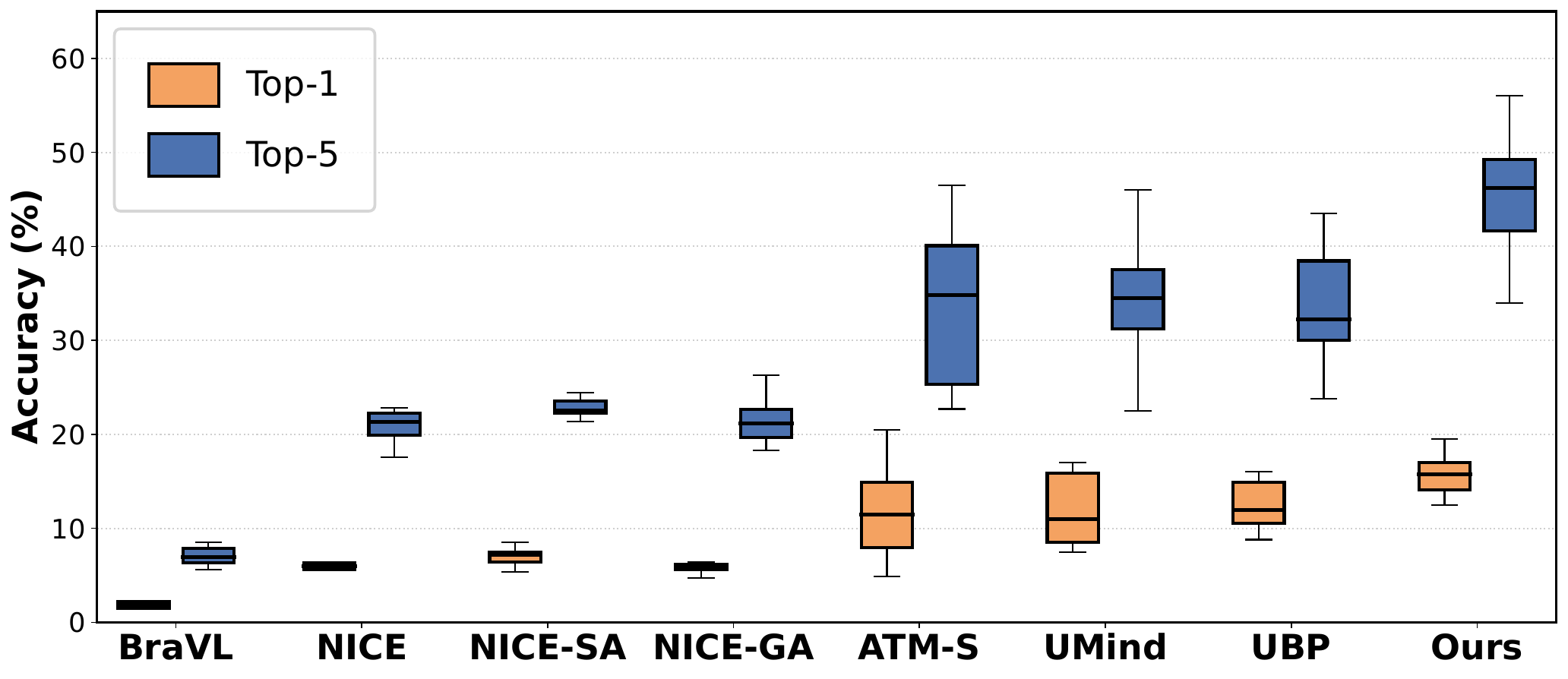}
    \caption{Cross-subject}
    \label{fig:boxplots_cross}
  \end{subfigure}
  \caption{Zero-shot decoding performance across within-subject and cross-subject settings. The proposed method achieves higher median accuracy and lower performance variability compared to prior methods, indicating improved robustness and generalization.}
  \label{fig:boxplots}
  \vspace{-6mm}
\end{figure}

In the within-subject setting, the proposed method achieves the best overall performance across subjects. 
As shown in Table~\ref{tab:zero_shot_200_subject_dependent}, our method obtains an average Top-1 accuracy of \textbf{54.8\%} and Top-5 accuracy of \textbf{85.6\%}, consistently outperforming all prior methods across subjects. 
Compared with the strongest baseline, UBP, our method improves Top-1 accuracy by \textbf{+3.9\%} and Top-5 accuracy by \textbf{+5.9\%}. 
These results indicate that the proposed framework effectively captures subject-specific neural dynamics while preserving strong semantic alignment with pretrained visual representations. 
As illustrated in Figure~\ref{fig:boxplots_within}, the proposed method also demonstrates higher median accuracy and lower variability across subjects, reflecting more stable and robust performance in the within-subject setting.


\begin{table*}[t]
\centering
\caption{Overall accuracy (\%) for 200-way zero-shot classification under the \textbf{within-subject} setting across 10 subjects. Our method achieves the best average performance.}
\label{tab:zero_shot_200_subject_dependent}

\small
\setlength{\tabcolsep}{4pt}
\renewcommand{\arraystretch}{1.05}
\resizebox{\textwidth}{!}{
\begin{tabular}{l*{22}{c}}
\toprule
\textbf{Method} &
\multicolumn{2}{c}{\textbf{Sub-01}} &
\multicolumn{2}{c}{\textbf{Sub-02}} &
\multicolumn{2}{c}{\textbf{Sub-03}} &
\multicolumn{2}{c}{\textbf{Sub-04}} &
\multicolumn{2}{c}{\textbf{Sub-05}} &
\multicolumn{2}{c}{\textbf{Sub-06}} &
\multicolumn{2}{c}{\textbf{Sub-07}} &
\multicolumn{2}{c}{\textbf{Sub-08}} &
\multicolumn{2}{c}{\textbf{Sub-09}} &
\multicolumn{2}{c}{\textbf{Sub-10}} &
\multicolumn{2}{c}{\textbf{Avg}} \\
\cmidrule(lr){2-3}\cmidrule(lr){4-5}\cmidrule(lr){6-7}
\cmidrule(lr){8-9}\cmidrule(lr){10-11}\cmidrule(lr){12-13}
\cmidrule(lr){14-15}\cmidrule(lr){16-17}\cmidrule(lr){18-19}
\cmidrule(lr){20-21}\cmidrule(lr){22-23}
& Top-1 & Top-5 & Top-1 & Top-5 & Top-1 & Top-5 & Top-1 & Top-5 & Top-1 & Top-5
& Top-1 & Top-5 & Top-1 & Top-5 & Top-1 & Top-5 & Top-1 & Top-5 & Top-1 & Top-5 & Top-1 & Top-5 \\
\midrule

BraVL \cite{du2023decoding}
& 6.1 & 17.9 & 4.9 & 14.9 & 5.6 & 15.1 & 6.0 & 13.4 & 6.0 & 18.2
& 6.5 & 20.4 & 8.8 & 23.7 & 4.3 & 14.0 & 7.0 & 19.7 & 7.0 & 19.7
& 5.8 & 17.5 \\

NICE \cite{songdecoding}
& 12.3 & 36.6 & 13.1 & 39.0 & 16.4 & 47.0 & 8.0 & 26.9 & 14.1 & 40.6
& 15.2 & 42.1 & 20.0 & 49.9 & 13.3 & 37.1 & 14.9 & 41.9 & 14.9 & 41.9
& 13.8 & 39.5 \\

NICE-SA \cite{songdecoding}
& 17.2 & 44.1 & 14.9 & 52.0 & 12.6 & 38.3 & 11.2 & 34.7 & 16.3 & 52.5
& 10.1 & 32.2 & 15.4 & 49.5 & 12.2 & 39.9 & 10.3 & 30.1 & 10.3 & 30.1
& 14.7 & 40.7 \\

NICE-GA \cite{songdecoding}
& 18.5 & 45.0 & 15.5 & 52.7 & 13.2 & 38.8 & 11.7 & 35.6 & 17.1 & 53.3
& 10.5 & 33.0 & 16.0 & 50.2 & 12.8 & 40.3 & 11.1 & 30.8 & 11.1 & 30.8
& 15.6 & 41.2 \\

ATM-S \cite{li2024visual}
& 25.6 & 60.4 & 22.0 & 54.5 & 25.0 & 62.4 & 31.4 & 60.9 & 12.9 & 43.0
& 21.3 & 51.1 & 30.5 & 61.5 & 38.8 & 72.0 & 34.4 & 51.5 & 29.1 & 63.5
& 28.5 & 60.4 \\

UMind \cite{xu2025umind}
& 27.0 & 56.0 & 32.0 & 70.0 & 34.0 & 70.0 & 36.0 & 70.5 & 23.0 & 50.5 
& 32.5 & 68.5 & 28.0 & 59.0 & 46.5 & 80.0 & 37.5 & 66.5 & 42.0 & 76.0 
& 33.9 & 66.7\\

CC \cite{zhang2025cognitioncapturer}
& 31.4 & 79.7 & 31.4 & 77.8 & 38.2 & 85.6 & 40.4 & 85.8 & 24.4 & 66.3
& 34.8 & 78.8 & 34.7 & 81.0 & 48.1 & 88.6 & 37.4 & 79.4 & 35.6 & 79.3
& 35.6 & 80.2 \\

UBP \cite{wu2025bridging}
& 41.2 & 70.5 & 51.2 & 80.9 & 51.2 & 82.0 & 51.1 & 76.9 & 42.2 & 72.8
& 57.5 & 83.5 & 49.0 & 79.9 & 58.6 & 85.8 & 45.1 & 76.2 & 61.5 & 88.2
& 50.9 & 79.7 \\

\midrule
\textbf{Ours}
& \textbf{53.0} & \textbf{84.5} & \textbf{51.0} & \textbf{84.0}
& \textbf{58.0} & \textbf{88.0} & \textbf{55.0} & \textbf{88.0}
& \textbf{38.0} & \textbf{69.5} & \textbf{54.5} & \textbf{87.0}
& \textbf{51.0} & \textbf{84.5} & \textbf{67.5} & \textbf{89.0}
& \textbf{52.5} & \textbf{87.5} & \textbf{67.5} & \textbf{94.0}
& \textbf{54.8} & \textbf{85.6} \\

\bottomrule
\end{tabular}
}
\vspace{-0.9mm}
\end{table*}

\begin{table*}[t]
\centering
\caption{Overall accuracy (\%) for 200-way zero-shot classification under the \textbf{cross-subject} setting (leave-one-subject-out). Our method achieves the best overall performance.}
\label{tab:zero_shot_200_sub_independent}

\small
\setlength{\tabcolsep}{3pt}
\renewcommand{\arraystretch}{0.95}

\resizebox{\textwidth}{!}{
\begin{tabular}{l*{22}{c}}
\toprule
\textbf{Method} &
\multicolumn{2}{c}{\textbf{Sub-01}} &
\multicolumn{2}{c}{\textbf{Sub-02}} &
\multicolumn{2}{c}{\textbf{Sub-03}} &
\multicolumn{2}{c}{\textbf{Sub-04}} &
\multicolumn{2}{c}{\textbf{Sub-05}} &
\multicolumn{2}{c}{\textbf{Sub-06}} &
\multicolumn{2}{c}{\textbf{Sub-07}} &
\multicolumn{2}{c}{\textbf{Sub-08}} &
\multicolumn{2}{c}{\textbf{Sub-09}} &
\multicolumn{2}{c}{\textbf{Sub-10}} &
\multicolumn{2}{c}{\textbf{Avg}} \\
\cmidrule(lr){2-3}\cmidrule(lr){4-5}\cmidrule(lr){6-7}
\cmidrule(lr){8-9}\cmidrule(lr){10-11}\cmidrule(lr){12-13}
\cmidrule(lr){14-15}\cmidrule(lr){16-17}\cmidrule(lr){18-19}
\cmidrule(lr){20-21}\cmidrule(lr){22-23}
& Top-1 & Top-5 & Top-1 & Top-5 & Top-1 & Top-5 & Top-1 & Top-5 & Top-1 & Top-5
& Top-1 & Top-5 & Top-1 & Top-5 & Top-1 & Top-5 & Top-1 & Top-5 & Top-1 & Top-5 & Top-1 & Top-5 \\
\midrule
BraVL \cite{du2023decoding}
& 2.3 & 8.0 & 1.5 & 6.3 & 1.4 & 5.9 & 1.7 & 6.7 & 1.5 & 5.6
& 1.8 & 7.2 & 2.1 & 8.1 & 2.2 & 7.6 & 1.6 & 6.4 & 2.3 & 8.5
& 1.8 & 7.0 \\

NICE \cite{songdecoding}
& 7.6 & 22.8 & 5.9 & 20.5 & 6.0 & 22.3 & 6.3 & 20.7 & 4.4 & 18.3
& 5.6 & 22.2 & 5.6 & 19.7 & 6.3 & 22.0 & 5.7 & 17.6 & 8.4 & 28.3
& 6.2 & 21.4 \\

NICE-SA \cite{songdecoding}
& 7.0 & 22.6 & 6.6 & 23.2 & 7.5 & 23.7 & 5.4 & 21.4 & 6.4 & 22.2
& 7.5 & 22.5 & 3.8 & 19.1 & 8.5 & 24.4 & 7.4 & 22.3 & 9.8 & 29.6
& 7.0 & 23.1 \\

NICE-GA \cite{songdecoding}
& 5.9 & 21.4 & 6.4 & 22.7 & 5.5 & 20.1 & 6.1 & 21.0 & 4.7 & 19.5
& 6.2 & 22.5 & 5.9 & 19.1 & 7.3 & 25.3 & 4.8 & 18.3 & 6.2 & 26.3
& 5.9 & 21.6 \\

ATM-S \cite{li2024visual}
& 10.5 & 26.8 & 7.1 & 24.8 & 11.9 & 33.8 & 14.7 & 39.4 & 7.0 & 23.9
& 11.1 & 35.8 & 16.1 & 43.5 & 15.0 & 40.3 & 4.9 & 22.7 & 20.5 & 46.5
& 11.8 & 33.7 \\

UMind \cite{xu2025umind}
& 7.5 & 31.0 & 7.5 & 32.0 & 15.5 & 37.5 & 17.0 & 38.5 & 8.0 & 22.5
& 10.5 & 37.5 & 11.5 & 30.0 & 16.0 & 46.0 & 10.0 & 32.5 & 16.5 & 36.5
& 12.0 & 34.4 \\

UBP \cite{wu2025bridging}
& 11.5 & 29.7 & 15.5 & 40.0 & 9.8 & 27.0 & 13.0 & 32.3 & 8.8 & 33.8
& 11.7 & 31.0 & 10.2 & 23.8 & 12.2 & 32.2 & 15.5 & 40.5 & 16.0 & 43.5
& 12.4 & 33.4 \\

\midrule
\textbf{Ours}
& \textbf{17.5} & \textbf{48.5} & \textbf{15.5} & \textbf{41.5}
& \textbf{9.0} & \textbf{34.0} & \textbf{12.5} & \textbf{37.5}
& \textbf{14.5} & \textbf{42.0} & \textbf{17.0} & \textbf{52.0}
& \textbf{14.0} & \textbf{45.0} & \textbf{16.0} & \textbf{47.5}
& \textbf{17.0} & \textbf{49.5} & \textbf{19.5} & \textbf{56.0}
& \textbf{15.3} & \textbf{45.4} \\
\bottomrule
\end{tabular}
}
\vspace{-0.9mm}
\end{table*}

\begin{table*}[t]
\centering
\caption{Overall accuracy (\%) for 200-way zero-shot classification under the \textbf{cross-session} setting. Our method achieves consistent improvements across subjects.}
\label{tab:zero_shot_200_cross_session}

\small
\setlength{\tabcolsep}{3pt}
\renewcommand{\arraystretch}{0.95}

\resizebox{\textwidth}{!}{
\begin{tabular}{l*{22}{c}}
\toprule
\textbf{Method} &
\multicolumn{2}{c}{\textbf{Sub-01}} &
\multicolumn{2}{c}{\textbf{Sub-02}} &
\multicolumn{2}{c}{\textbf{Sub-03}} &
\multicolumn{2}{c}{\textbf{Sub-04}} &
\multicolumn{2}{c}{\textbf{Sub-05}} &
\multicolumn{2}{c}{\textbf{Sub-06}} &
\multicolumn{2}{c}{\textbf{Sub-07}} &
\multicolumn{2}{c}{\textbf{Sub-08}} &
\multicolumn{2}{c}{\textbf{Sub-09}} &
\multicolumn{2}{c}{\textbf{Sub-10}} &
\multicolumn{2}{c}{\textbf{Avg}}\\
\cmidrule(lr){2-3}\cmidrule(lr){4-5}\cmidrule(lr){6-7}
\cmidrule(lr){8-9}\cmidrule(lr){10-11}\cmidrule(lr){12-13}
\cmidrule(lr){14-15}\cmidrule(lr){16-17}\cmidrule(lr){18-19}
\cmidrule(lr){20-21}\cmidrule(lr){22-23}
& Top-1 & Top-5 & Top-1 & Top-5 & Top-1 & Top-5 & Top-1 & Top-5 & Top-1 & Top-5
& Top-1 & Top-5 & Top-1 & Top-5 & Top-1 & Top-5 & Top-1 & Top-5 & Top-1 & Top-5 & Top-1 & Top-5 \\
\midrule

UBP
& 28.5 & 59.5 & \textbf{39.0} & 67.0 & 42.5 & 73.5 & 37.0 & 71.5 & 36.0 & 65.5 
& \textbf{44.0} & 76.5 & 36.0 & 69.0 & 47.0 & 79.5 & 37.5 & 67.0 & 49.5 & 80.5
& 39.7 & 71.0 \\

\textbf{Ours}
& \textbf{31.0} & \textbf{67.5} & 38.5 & \textbf{76.4} & \textbf{44.0} & \textbf{81.1} & \textbf{38.6} & \textbf{75.7} & \textbf{37.0} & \textbf{75.0} 
& 40.0 & \textbf{78.2} & \textbf{39.0} & \textbf{77.0} & \textbf{50.2} & \textbf{86.6} & \textbf{39.3} & \textbf{77.5} & \textbf{50.6} & \textbf{85.2}
& \textbf{40.8} & \textbf{78.0} \\

\bottomrule
\end{tabular}
}
\vspace{-3mm}
\end{table*}

\begin{table*}[!t]
\centering

\caption{
Ablation study of temporal, spectral, and spatial EEG representation modules under the within-subject setting on THINGS-EEG benchmark.
Results are reported in Top-1 and Top-5 classification accuracy (\%) for Subject 10.
}
\label{tab:module_ablation}
\resizebox{\textwidth}{!}{
\small
\begin{tabular}{lccccccc}
\toprule
\textbf{Metric} 
& \textbf{Full Model} 
& \textbf{w/o Temporal} 
& \textbf{w/o Spatial} 
& \textbf{w/o Spectral} 
& \textbf{Temporal Only} 
& \textbf{Spatial Only} 
& \textbf{Spectral Only} \\
\midrule
\textbf{Top-1} $\uparrow$ 
& \textbf{67.5} & 63.5 & 60.0 & 59.0 & 57.5 & 49.5 & 54.0 \\
\textbf{Top-5} $\uparrow$ 
& \textbf{94.0} & 92.0 & 90.0 & 92.0 & 86.0 & 83.0 & 89.0 \\
\bottomrule
\end{tabular}
}
\vspace{-3mm}
\end{table*}

\begin{table*}[!t]
\centering
\caption{Ablation study evaluating the contribution of CPCA and RCR regularization terms under a within-subject setting.
}
\label{tab:ablation_main}

\small
\setlength{\tabcolsep}{4pt}
\renewcommand{\arraystretch}{1.05}
\resizebox{\textwidth}{!}{
\begin{tabular}{l*{22}{c}}
\toprule
\textbf{Configuration} &
\multicolumn{2}{c}{\textbf{Sub-01}} &
\multicolumn{2}{c}{\textbf{Sub-02}} &
\multicolumn{2}{c}{\textbf{Sub-03}} &
\multicolumn{2}{c}{\textbf{Sub-04}} &
\multicolumn{2}{c}{\textbf{Sub-05}} &
\multicolumn{2}{c}{\textbf{Sub-06}} &
\multicolumn{2}{c}{\textbf{Sub-07}} &
\multicolumn{2}{c}{\textbf{Sub-08}} &
\multicolumn{2}{c}{\textbf{Sub-09}} &
\multicolumn{2}{c}{\textbf{Sub-10}} &
\multicolumn{2}{c}{\textbf{Avg}} \\
\cmidrule(lr){2-3}\cmidrule(lr){4-5}\cmidrule(lr){6-7}
\cmidrule(lr){8-9}\cmidrule(lr){10-11}\cmidrule(lr){12-13}
\cmidrule(lr){14-15}\cmidrule(lr){16-17}\cmidrule(lr){18-19}
\cmidrule(lr){20-21}\cmidrule(lr){22-23}
& Top-1 & Top-5 & Top-1 & Top-5 & Top-1 & Top-5 & Top-1 & Top-5 & Top-1 & Top-5
& Top-1 & Top-5 & Top-1 & Top-5 & Top-1 & Top-5 & Top-1 & Top-5 & Top-1 & Top-5 & Top-1 & Top-5 \\
\midrule

Full model

& 53.0 & 84.5 
& 51.0 & 84.0 
& 58.0 & 88.0 
& 55.0 & 88.0 
& 38.0 & 69.5 
& 54.5 & 87.0 
& 51.0 & 84.5 
& 67.5 & 89.0 
& 52.5 & 87.5 
& 67.5 & 94.0 
& 54.8 & 85.6 \\

w/o CPCA
& 53.5 & 82.5 
& 49.5 & 81.5 
& 59.5 & 89.5 
& 54.5 & 87.0 
& 36.5 & 70.0 
& 53.5 & 86.0 
& 50.5 & 86.0 
& 66.0 & 89.0 
& 52.0 & 86.0 
& 70.5 & 93.0 
& 54.6 & 85.1 \\

w/o RCR
& 55.0 & 80.5 
& 49.0 & 82.0 
& 40.0 & 78.5 
& 52.0 & 87.5 
& 43.0 & 71.0 
& 50.0 & 87.0 
& 53.5 & 88.0 
& 66.5 & 91.5 
& 52.0 & 85.5 
& 67.5 & 95.0 
& 52.9 & 84.7 \\

w/o CPCA \& RCR
& 39.0 & 73.5 
& 33.0 & 67.5 
& 46.0 & 82.5 
& 44.5 & 78.5 
& 22.5 & 53.5 
& 40.5 & 75.5 
& 43.5 & 81.5 
& 58.5 & 88.5 
& 37.0 & 66.0 
& 54.5 & 85.0 
& 41.9 & 75.2 \\

\bottomrule
\end{tabular}
}
\vspace{-6mm}
\end{table*}


\subsection{Cross-Subject zero-shot classification}
In the cross-subject setting, we evaluate whether the proposed method can generalize to unseen subjects despite substantial inter-subject variability in EEG signals. 
The model is trained on all but one subject and tested on the held-out subject. 
As shown in Table~\ref{tab:zero_shot_200_sub_independent}, the proposed method achieves an average Top-1 accuracy of \textbf{15.3\%} and Top-5 accuracy of \textbf{45.4\%}. 
Compared with UBP, the strongest baseline in this setting, our method improves Top-1 accuracy by \textbf{+2.9\%} and Top-5 accuracy by \textbf{+12.0\%}. 
These results demonstrate stronger generalization capability under subject-level distribution shifts. 
Figure~\ref{fig:boxplots_cross} further shows higher median accuracy and lower performance variability across subjects, indicating more stable cross-subject decoding performance. 
Overall, these results suggest that the proposed method learns robust and semantically meaningful subject-invariant representations.

\subsection{Cross-Session zero-shot classification}

The cross-session setting evaluates whether the proposed method remains effective when training and testing are conducted on different recording sessions of the same subject. This setting is more challenging because EEG signals can exhibit session-dependent distribution shifts caused by recording noise, fatigue, electrode variability, and changes in subject state. To the best of our knowledge, this is the first systematic evaluation to investigate zero-shot EEG--image semantic decoding in the cross-session setting. Since most existing methods do not report results under this protocol, we compare with UBP~\cite{wu2025bridging}, a strong recent baseline. As shown in Table~\ref{tab:zero_shot_200_cross_session}, the proposed method consistently outperforms UBP across subjects, demonstrating improved robustness and stronger cross-session generalization through stable and semantically meaningful EEG representations.

\subsection{Ablation Study}
\label{sec:ablation}

We conduct ablation studies to assess the contribution of the main architectural modules and regularization terms. 
All ablations follow the same training protocol and evaluation metrics as the full model. 
Table~\ref{tab:module_ablation} reports the within-subject results when removing the temporal, spectral, or spatial module individually, showing the contribution of each EEG representation branch.

We also evaluate the effect of the proposed regularization terms by removing CPCA and RCR individually and jointly. 
Table~\ref{tab:ablation_main} reports the average Top-1 and Top-5 accuracy across all 10 subjects under within-subject and cross-subject settings. 
Removing either regularizer degrades performance, while removing both causes the largest drop. 
The larger degradation from removing RCR suggests that relational consistency regularization is particularly important for preserving discriminative EEG representations. 

\begin{figure*}[t]
\centering

\begin{subfigure}[t]{0.185\linewidth}
    \centering
    \includegraphics[width=\linewidth]{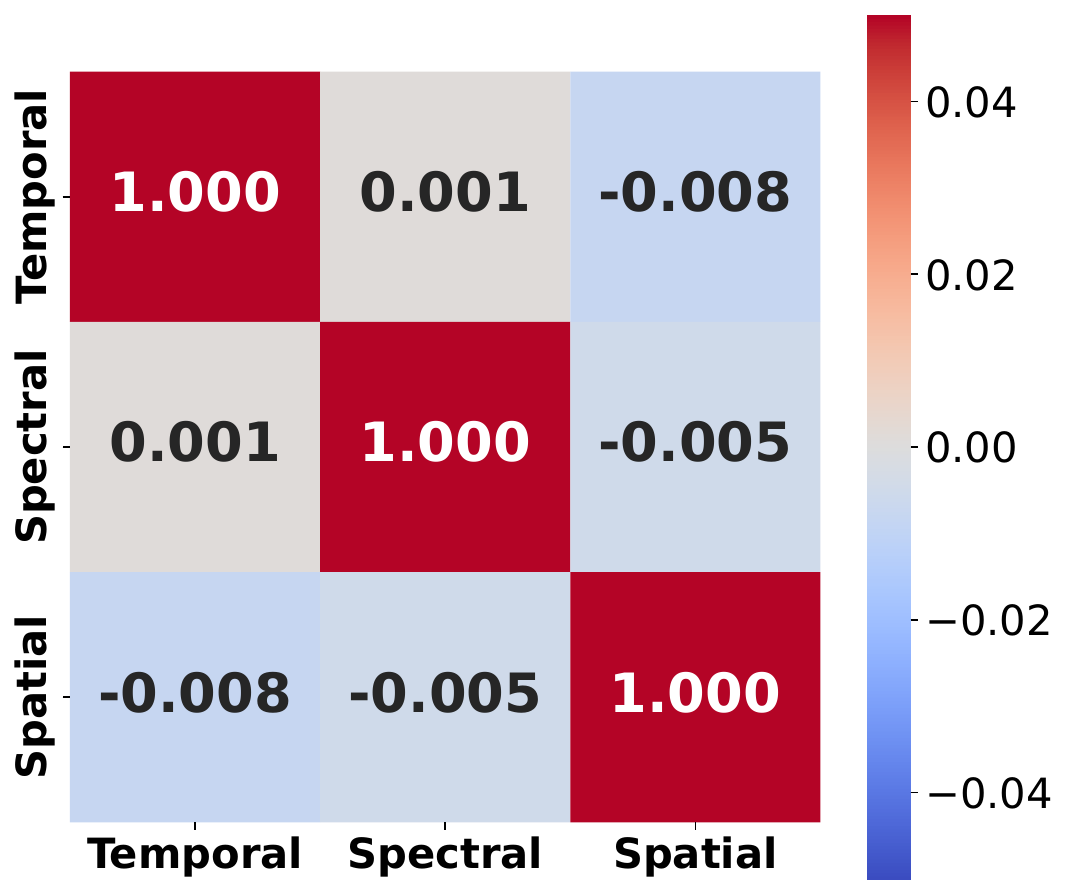}
    \caption{Cosine similarity}
    \label{fig:cos_top1}
\end{subfigure}
\hfill
\begin{subfigure}[t]{0.185\linewidth}
    \centering
    \includegraphics[width=\linewidth]{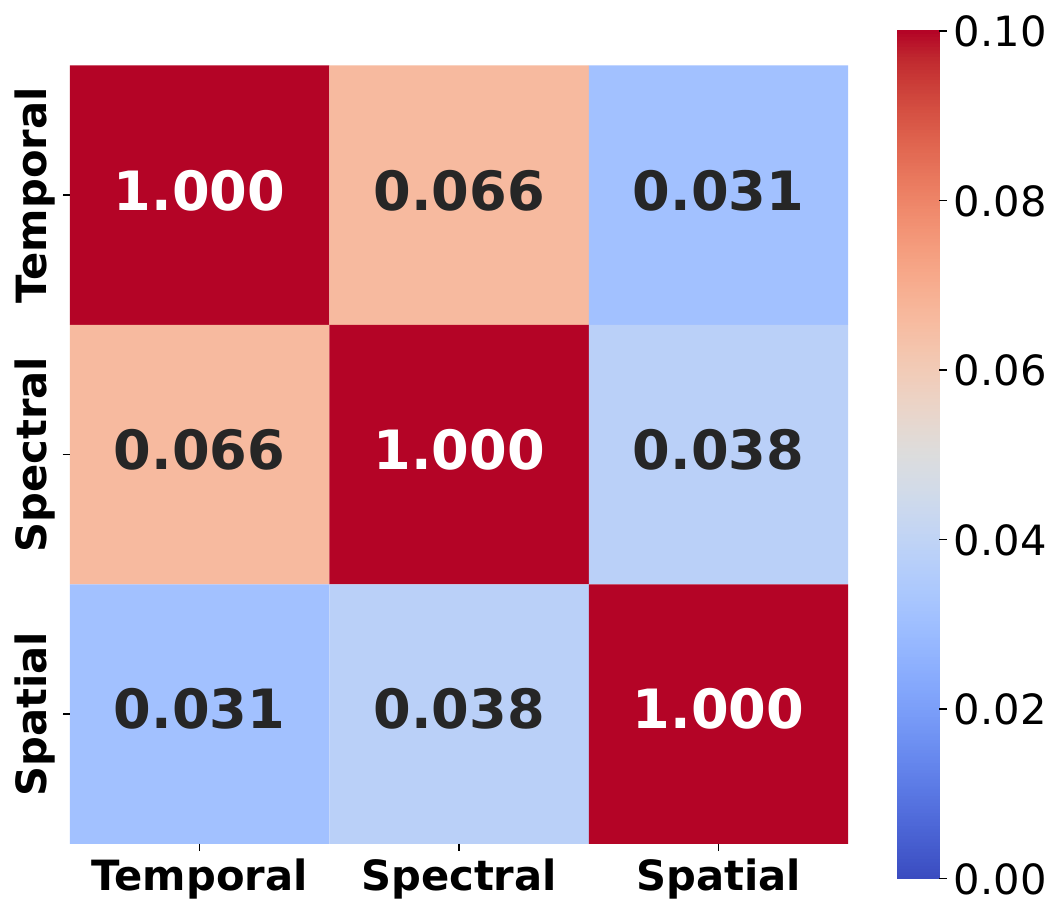}
    \caption{CKA similarity}
    \label{fig:cka_top1}
\end{subfigure}
\hfill
\begin{subfigure}[t]{0.185\linewidth}
    \centering
    \includegraphics[width=\linewidth]{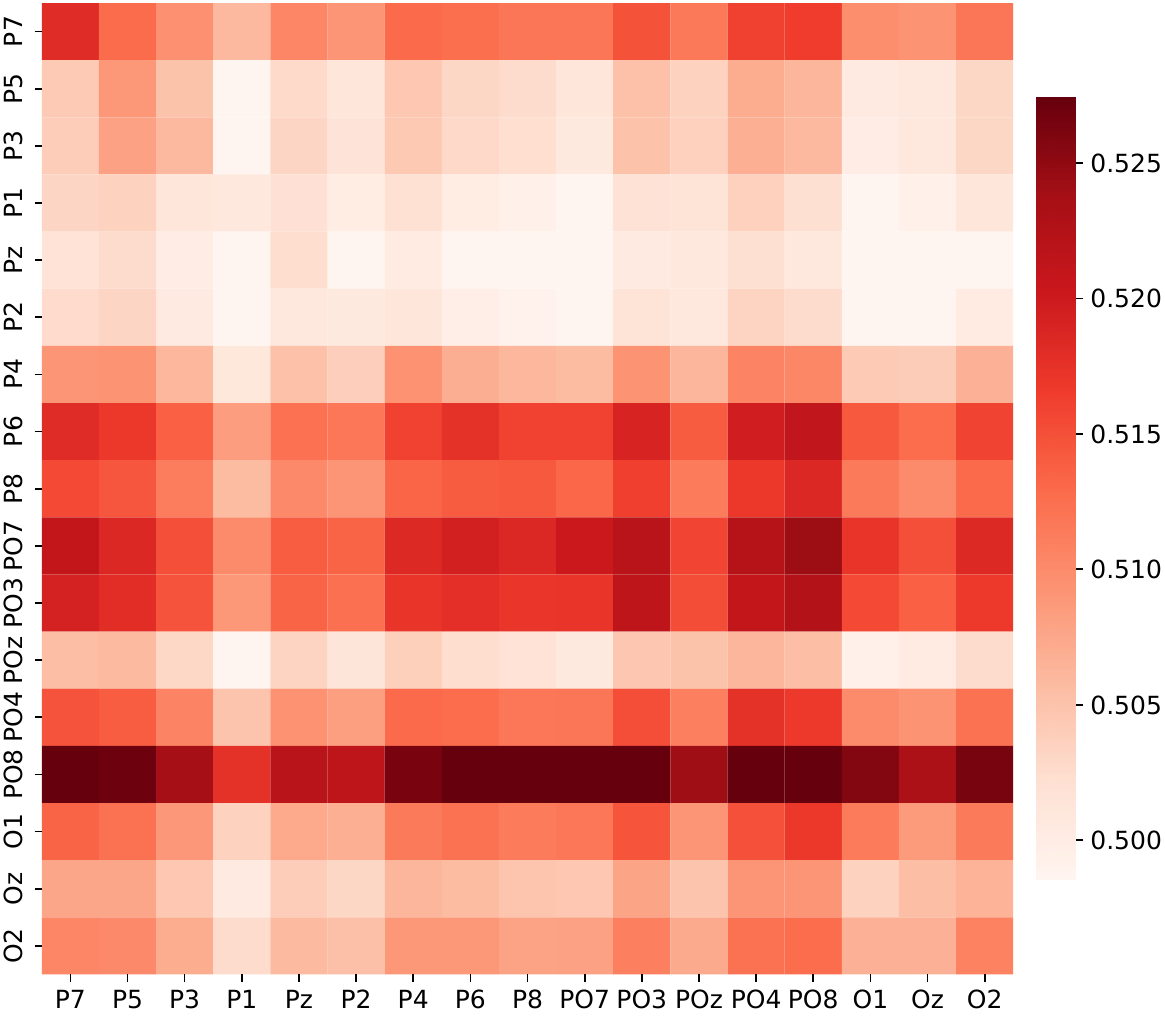}
    \caption{Connectivity heatmap}
    \label{fig:gcn_connectivity_heatmap}
\end{subfigure}
\hfill
\begin{subfigure}[t]{0.185\linewidth}
    \centering
    \includegraphics[width=\linewidth]{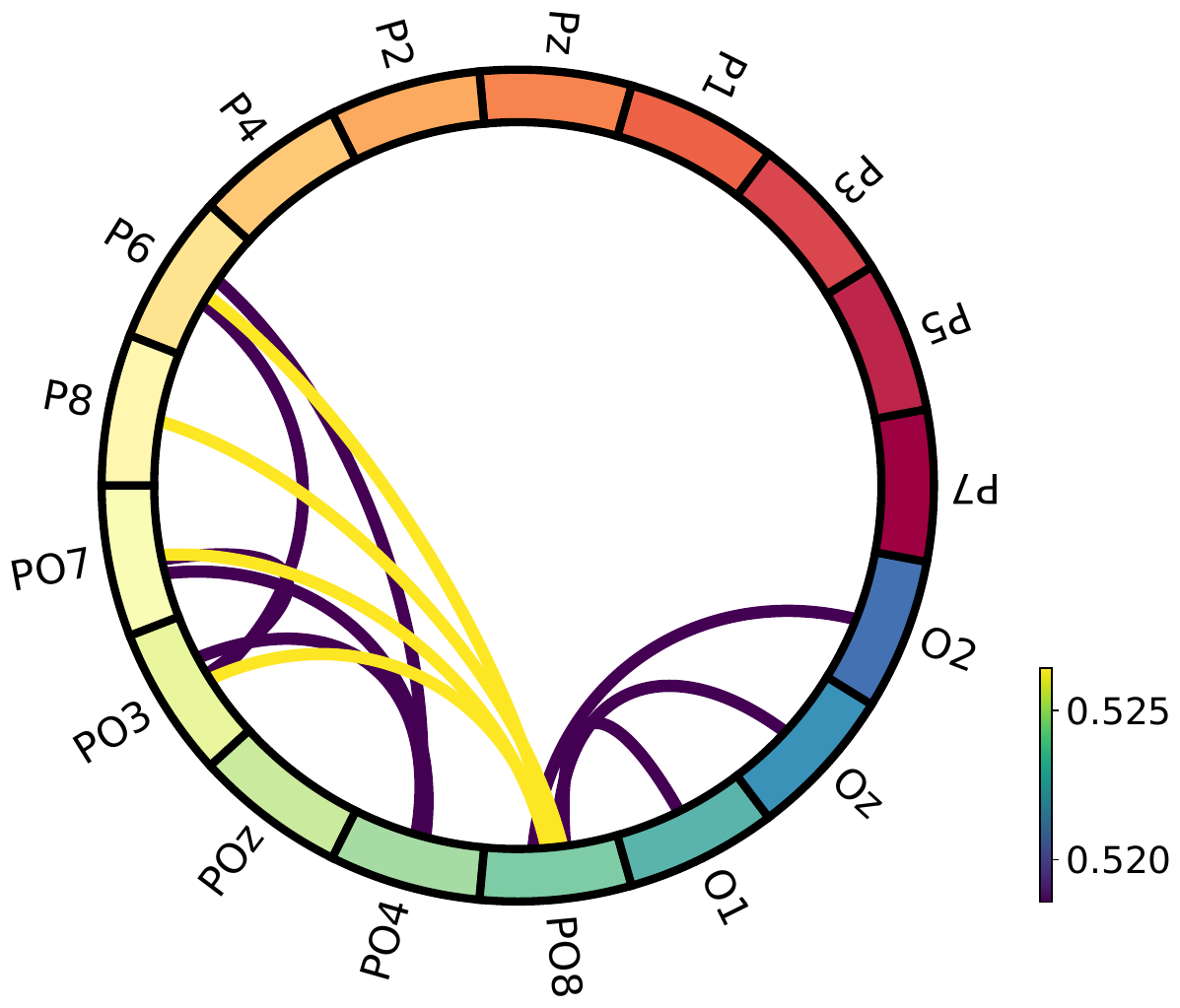}
    \caption{Top graph connections}
    \label{fig:gcn_connectivity_circle}
\end{subfigure}
\hfill
\begin{subfigure}[t]{0.185\linewidth}
    \centering
    \includegraphics[width=\linewidth]{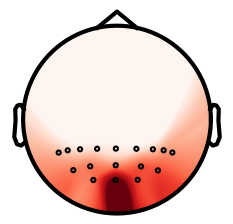}
    \caption{Mean topography}
    \label{fig:Mean_Topomap}
\end{subfigure}


\caption{
Analysis of learned EEG representations and spatial connectivity.
(a) and (b) show inter-view similarity among temporal, spectral, and spatial representations using cosine similarity and CKA.
Low off-diagonal values indicate complementary multiview representations.
(c) shows the learned adaptive GCN adjacency matrix averaged across subjects.
(d) visualizes the strongest learned graph connections.
(e) shows the averaged electrode-importance topomap.
The connectivity and topographic patterns emphasize posterior and occipital regions associated with visual processing.
}
\label{fig:representation_analysis}
\vspace{-0.7em}
\end{figure*}

\section{Discussion}

Beyond the main decoding results, we analyze the learned representations to interpret how the proposed framework supports EEG--visual alignment. 
Specifically, we examine inter-view similarity among temporal, spectral, and spatial branches, and visualize the spatial connectivity patterns learned by the adaptive GCN. 
Limitations and broader impacts, including generalization beyond controlled EEG settings and neural-data privacy considerations.

\subsection{Representation disentanglement}

We analyze whether the temporal, spectral, and spatial branches capture complementary information by measuring inter-view similarity using cosine similarity and Centered Kernel Alignment (CKA). 
As shown in Figure~\ref{fig:representation_analysis}(a) and (b), cosine similarities remain close to zero ($|\text{mean}| \leq 0.01$), while CKA scores are generally below $0.07$. 
These low similarities indicate that the learned views are largely non-redundant. 
The slightly higher similarity between temporal and spectral views is expected, since frequency-domain patterns are derived from temporal EEG dynamics, whereas spatial features remain more distinct. 

\subsection{Learned EEG connectivity}

We further examine the adaptive GCN by visualizing the learned adjacency matrices averaged across subjects. 
Figure~\ref{fig:representation_analysis}(c)--(e) shows that the learned connectivity is non-uniform and consistently emphasizes posterior and occipital electrodes, including PO7, PO8, O1, Oz, and neighboring parieto-occipital regions. 
The sparse graph also reveals long-range interactions between parietal and occipital areas. 
These patterns suggest that the graph-learning module captures spatial dependencies relevant to visual EEG decoding rather than relying on random or uniformly distributed electrode interactions. 

\section{Conclusion}

In this work, we presented a unified framework for EEG-based zero-shot visual decoding through multiview neural representation learning. By jointly modeling temporal dynamics, adaptive spectral structure, and spatial electrode interactions, the proposed method learns compact EEG embeddings that align effectively with pretrained visual representations in a shared semantic space. Experiments on THINGS-EEG benchmark show consistent improvements over prior approaches under within-subject, cross-subject, and cross-session settings. In particular, our systematic cross-session evaluation highlights the importance of structured EEG inductive biases for robust generalization under realistic recording variability. These findings suggest that multiview EEG modeling provides a principled foundation for brain--visual alignment and offers a scalable path toward more reliable EEG-based neural decoding. 



\bibliographystyle{unsrtnat}
\bibliography{example_paper}

\end{document}